\documentclass[11pt]{article}

\PassOptionsToPackage{table}{xcolor}

\usepackage{acl}

\usepackage{times}
\usepackage{latexsym}
\usepackage[T1]{fontenc}
\usepackage[utf8]{inputenc}
\usepackage{microtype}
\usepackage{inconsolata}
\usepackage{graphicx}
\usepackage{booktabs}
\usepackage{amsmath}
\usepackage{amssymb}
\usepackage{enumitem}
\usepackage{textcomp}
\usepackage{pifont}
\usepackage{url}

\DeclareUnicodeCharacter{266B}{\textmusicalnote}

\graphicspath{{figures/}{../}}

\newcommand{\ours}{\textsc{Skill2Steer}}
\newcommand{\sig}{\ensuremath{\sigma}}
\newcommand{\hdown}{\ensuremath{\mathbf{W}^{\text{down}}}}
\newcommand{\hup}{\ensuremath{\mathbf{W}^{\text{up}}}}
\newcommand{\yes}{\ding{51}}
\newcommand{\no}{\ding{55}}

\title{\ours{}: One-Pass Hypernetwork for Mapping Natural-Language Skills
to Low-Rank Post-Block Activation Shifts}

\author{
	Ran Li\textsuperscript{1} \quad
	Lei Chen\textsuperscript{1,2} \\
	\textsuperscript{1}Hong Kong University of Science and Technology  \\
	\textsuperscript{2}Hong Kong University of Science and Technology(Guangzhou)  \\
	\texttt{rlibb@connect.ust.hk} \quad \texttt{leichen@cse.ust.hk}
}

\begin{document}
\maketitle

\begin{abstract}
Compiling a natural-language agent skill into model parameters removes the
skill text from the prompt and leaves behind a small artifact that can be
distributed and reused. Existing methods differ in where they inject the
compiled quantity, and all of them assume that the artifact carries the content
of the skill. We test that assumption. Through three controls on a retrieval
agent, we find that a compiled skill shift carries a domain-general behavioural
prior, not the content of the skill. A wrong skill document performs on par with
the correct one, a single averaged bundle matches per-skill routing under a
two-sided equivalence test, and a same-norm random direction collapses to the
no-injection baseline, which shows that the trained direction stays
load-bearing. We then introduce \ours{}, a one-pass hypernetwork that reads a
skill document once and maps it to low-rank activation shifts added at the
output of every decoder block. Each skill compiles to 2.36M parameters, 9.25
times fewer than a weight-space adapter, while the base weights stay frozen and
the generator runs only at compile time. The shift can be scheduled across
retrieval steps and deployed as one routing-free bundle. On SearchQA, a
seven-dataset benchmark, \ours{} reaches 37.9 exact match,
9.8 points above the frozen model and 2.3 points above a weight-space method
that ships 9.25 times more parameters per skill. Code and compiled bundles are
available at \url{https://github.com/rlisml/skill2steer}.
\end{abstract}

\section{Introduction}

Language agents increasingly act through natural-language skill documents,
short texts that describe how to search, when to stop, and how to answer. A
recent line of work compiles such a document into model parameters, so that
inference runs without the skill text and each skill becomes a small artifact
that can be shipped and reused. LatentSkill \citep{yu2026latentskill} distils a
textual skill into in-weight latent skills, and SHINE \citep{liu2026shine}
generates LoRA weights from a context in a single pass. The compiled quantity
may be a weight update or an activation offset, and both families report
competitive accuracy on retrieval tasks.

\begin{figure}[t]
\centering
\includegraphics[width=0.51\linewidth]{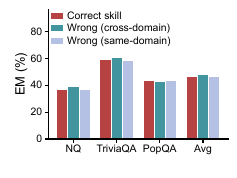}\hfill
\includegraphics[width=0.49\linewidth]{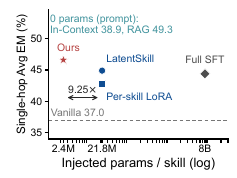}
\caption{Compiled skill shifts are domain-general and cheap. A wrong skill
document matches the correct one, and the activation-space shift sits at the
small-parameter end of the accuracy frontier.}
\label{fig:motivation}
\end{figure}

These methods share an assumption that has not been tested. The artifact is
taken to carry the content of the skill, which is why the standard recipe keeps
one artifact per skill and routes each query to the matching one. We ask a more
basic question. When a hypernetwork compiles a skill into parameters, what does
the injection actually carry?

Three controls answer the question for a retrieval agent. Replacing the skill
document with a wrong one leaves accuracy unchanged
(Figure~\ref{fig:motivation}, left). Averaging several compiled
artifacts into a single bundle matches per-skill routing under a two-sided
equivalence test. Removing the trained component and substituting a random
direction of the same norm drops the model back to its no-injection accuracy.
The first two results point to a domain-general behavioural prior as the
carrier. The third shows that the trained direction still matters, so the prior
has to be reached from the training distribution. A skill document then acts as
a key that triggers the prior, and the teeth of the key matter less than the
distribution it is drawn from.

We build \ours{} on this finding. A one-pass hypernetwork reads the skill
document once, through a frozen encoder, and emits low-rank factors for every
decoder block. The factors are added to the residual stream at the block
output, which keeps the base weights untouched and matches an injection site
that inference engines already expose for steering \citep{adila2026weight,li2026latentharnesscompilingagent}. Each
skill costs 2.36M parameters, 9.25 times fewer than a weight-space adapter, and
the generator runs only at compile time, so it never enters the deployment
budget.

On SearchQA, a seven-dataset retrieval benchmark with 3125 questions,
\ours{} reaches 37.9 exact match. That is 9.8 points above the frozen model and
2.3 points above a weight-space method that ships 9.25 times more parameters per
skill. The gain is a behavioural one. The shift collapses a long, degenerate
completion into a short search-then-answer action, and on multi-hop questions it
can be removed after the first retrieval step to recover further points. Because
the carrier is domain-general, the same shift can be deployed as a single
bundle with no routing.

Our contributions are as follows.

\begin{itemize}[leftmargin=1.2em,itemsep=2pt,topsep=2pt,parsep=0pt]
\item \textbf{\ours{}.} A one-pass hypernetwork that maps a natural-language
skill to low-rank activation shifts at the output of every decoder block. The
artifact holds 2.36M parameters per skill, 9.25 times fewer than weight-space
compilation, and leaves the base weights frozen.
\item \textbf{A behaviour-prior account of skill injection.} Evidence that the
compiled shift carries a domain-general behavioural prior. Wrong skill documents
and a single universal bundle are equivalent to the correct per-skill artifact,
while a same-norm random direction is not.
\item \textbf{Schedulability and routing-free deployment.} A criterion that
predicts when the shift should be withdrawn during a trajectory, together with a
recipe in which one bundle covers every skill at no measurable cost.
\end{itemize}

\section{Related Work}

\paragraph{Skill injection for language agents.}
A skill document tells an agent how to act. One family of methods keeps the
document in context, either as a retrieved passage or as an in-context
instruction, which leaves the base model untouched but pays a per-query context
cost \citep{li2026skillsinjector}. A second family distils the skill into
parameters. SHINE \citep{liu2026shine} and Text-to-LoRA \citep{charakorn2025text}
generate LoRA weights from a context in a single pass, Doc-to-LoRA
\citep{charakorn2026doc} extends this to long documents, and LatentSkill
\citep{yu2026latentskill} maps a textual skill to in-weight latent skills.
Skill-to-LoRA \citep{zhang2026skill}, Parametric Skills \citep{zhao2026parametric},
and Skill0 \citep{lu2026skill0} pursue the same goal from the behaviour side. A
third family edits internal representations at inference, through steering
vectors \citep{turner2023steering,rimsky2024steering}, representation finetuning
\citep{wu2024reft}, or hypernetwork-generated steering at scale
\citep{sun2025hypersteer,sharma2026cold}. \ours{} sits in the third family. It
conditions on a skill document, injects at every block output, and, unlike prior
work, reports what the injected quantity carries.

\paragraph{Hypernetworks for adaptation.}
Hypernetworks generate the parameters of a target network from a conditioning
input \citep{ye2021learning,shenaj2025lora}. Recent instantiations map a task
description or a context to adapter weights \citep{charakorn2025text,liu2026shine,tan2026instant}
or to a steering vector \citep{sun2025hypersteer}. Our generator follows this
recipe but emits a low-rank factor pair for each of the $L$ decoder blocks, and
it is trained once across a corpus of agent skills and is not retrained per task.

\paragraph{Activation-space intervention.}
Adding a vector or a low-rank term to the residual stream is an established way
to control a frozen model \citep{turner2023steering,zou2023representation,li2023inference}.
Contrastive activation addition \citep{rimsky2024steering} and function vectors
\citep{todd2024function} locate such directions from data, and ReFT
\citep{wu2024reft} learns low-rank edits inside a block. A recent analysis shows
that a weight update can be rewritten as an activation shift and that a
post-block site gives a stable intervention point \citep{adila2026weight,li2026learningdecidereasonparameterefficient}. We
take that site and let a hypernetwork choose the shift from a skill document.

\section{\ours{}}

\begin{figure*}[t]
\centering
\includegraphics[width=0.99\textwidth]{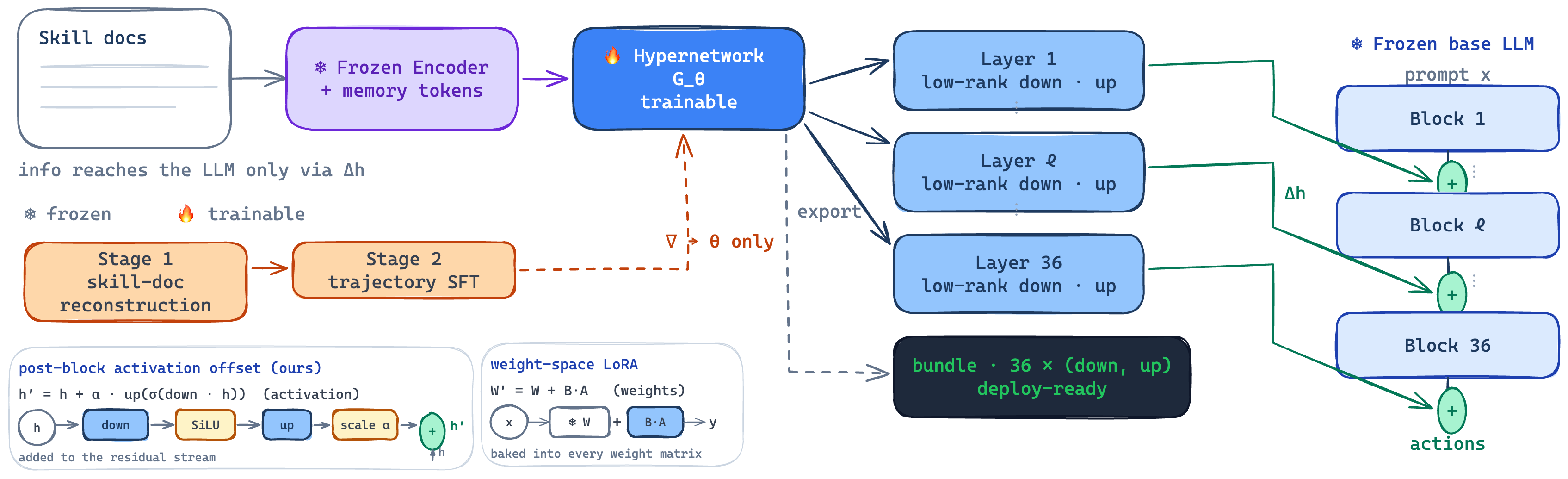}
\caption{\ours{}. A frozen encoder reads the skill document once, a
hypernetwork emits one low-rank factor pair per decoder block, and the offset is
added at each block output. The exported bundle mounts on the inference engine
with no change to the base weights.}
\label{fig:overview}
\end{figure*}

The method has three parts. A frozen skill encoder compresses the document into
a fixed-length summary. A trainable hypernetwork turns that summary into
per-layer low-rank factors. A parameter-free module adds the resulting offset at
each block output. Figure~\ref{fig:overview} shows the pipeline.

\subsection{Problem setup}

Let $M_{\varphi}$ be a frozen decoder-only transformer with $L$ layers and
hidden width $H$, instantiated here as Qwen3-8B with $L=36$ and $H=4096$
\citep{yang2025qwen3}. The parameters $\varphi$ are not updated at any point.
Write $\mathrm{Block}_l$ for the $l$-th decoder block and $h^{(l)}$ for its
output, the post-block residual stream. Given a skill document $s$, we train a
compiler $\mathcal{G}_{\theta}$ so that $M_{\varphi}$, with $\varphi$ fixed,
behaves as if it had already acquired $s$. The skill text reaches the model only
through the quantity $\mathcal{G}_{\theta}(s)$, and never enters the prompt at
inference time.

\subsection{Skill encoder}

The skill document is tokenised by the frozen tokenizer. To reduce the whole
document to a fixed-length summary, we append $N=2r=16$ learnable memory tokens
and run one forward pass through the frozen base model under \texttt{no\_grad}.
The last $N$ hidden states of every layer form the memory states
$M \in \mathbb{R}^{L \times N \times H}$, where $M_l \in \mathbb{R}^{N \times H}$
is the summary of $s$ at layer $l$. The rank $r$ defaults to 8, and the factor
$2r$ lets the summary be split into a down and an up part later.

\subsection{Hypernetwork generator}

The generator $\mathcal{G}_{\theta}$ is a four-layer transformer encoder with
width 4096, 32 heads, and an 8192-wide feed-forward network. It mixes the
$L \times N$ memory tokens across layers and across positions, and returns one
low-rank factor pair per layer,
\begin{equation}
\begin{aligned}
\{(\hdown_l, \hup_l)\}_{l=1}^{L} &= \mathrm{split}\big(\mathcal{G}_{\theta}(M)\big), \\
\hdown_l &\in \mathbb{R}^{r \times H}, \quad \hup_l \in \mathbb{R}^{H \times r}.
\end{aligned}
\end{equation}
Each layer contributes $2rH = 65536$ scalars, so a skill is represented by
$L \times 2 \times r \times H = 2{,}359{,}296 \approx 2.36$M scalars. The
generator itself holds about 537.3M parameters, but it runs once per skill at
compile time and is not part of the deployed artifact.

\subsection{Post-block injection}

At the output of layer $l$ the offset is added to every token position,
\begin{equation}
\Delta h_l = \alpha \, \hup_l \sig\!\big(\hdown_l \, h^{(l)}\big),
\; 
\tilde{h}^{(l)} = h^{(l)} + \Delta h_l ,
\end{equation}
where $\sig$ is the SiLU nonlinearity and $\alpha$ is an injection strength set
to $10^{-2}$ during training. The initial relative perturbation
$\lVert \Delta h \rVert / \lVert h \rVert$ stays between 0.9\% and 5.7\%. The
offset $\tilde{h}^{(l)}$ feeds the next block. This form matches a standard
post-block low-rank adapter, so a trained product can be exported to an engine
that already supports low-rank steering.

\subsection{Two-stage training}

Only $\theta$ is updated, which includes the memory tokens and the generator,
while the base model stays frozen. Stage one performs skill-document
pretraining on roughly 170K agent skill documents. Each document is used under
two self-supervised objectives, a reconstruction objective whose evidence is the
full document and a completion objective whose evidence is a random prefix. Both
targets are the complete skill text, and the skill text enters only the encoder,
so the information has to travel through the generated offset. Stage two adds
trajectory supervision on about 8500 pairs of a skill document and an expert
action trace. The loss is the cross-entropy on expert action tokens. Stage two
starts from the stage-one checkpoint and uses 2400 micro-steps on a single A800,
which is about one ninth of the reference training compute.

\subsection{Export and inference}

The trained strength $\alpha$ is folded into $\hup$ at export time, and the
bundle manifest records a unit strength with an explicit \texttt{scale} scalar
that can be varied at inference. The per-layer factors are mounted through the
native steering interface of the engine, and neither the base model nor the
engine is modified. One bundle corresponds to one skill. Because the carrier
turns out to be domain-general, the bundle can also be averaged across skills
and shipped as a single artifact, which we test in Section~\ref{sec:inject}.

\section{Experimental Setup}

\paragraph{Task.}
We evaluate on SearchQA, an interactive retrieval question-answering task. Given
a question, an agent interacts with a retriever for at most $K=4$ steps and
chooses at each step between issuing a search query and giving a final answer.
The retriever is E5 \citep{wang2024multilingual} over the wiki-18 corpus of
about 21M passages, and each step returns the top three passages. The metric is
exact match (EM) after answer normalisation. Decoding is greedy with a limit of
2048 new tokens and reasoning traces disabled.

\paragraph{Data.}
The evaluation set holds 3125 questions over seven datasets. The single-hop
group has 1500 questions from Natural Questions
\citep{kwiatkowski2019natural}, TriviaQA \citep{joshi2017triviaqa}, and PopQA
\citep{mallen2023not}, each with 500 questions and routed to a direct retrieval
skill. The multi-hop group has 1625 questions from HotpotQA
\citep{yang2018hotpotqa}, 2WikiMultihopQA \citep{ho2020constructing}, and
MuSiQue \citep{trivedi2022musique}, each with 500 questions and routed to a
multi-hop or comparison skill, plus 125 questions from Bamboogle
\citep{press2023measuring}. Datasets that appear in the agent fine-tuning corpus
are marked in-domain, and the remaining ones are unseen during training.

\paragraph{Comparisons.}
We report the frozen base model, four prompting and retrieval baselines, and
seven skill-compilation methods drawn from prior work, including the
weight-space methods SHINE \citep{liu2026shine}, Text-to-LoRA
\citep{charakorn2025text}, and LatentSkill \citep{yu2026latentskill}. All
skill-compilation methods share the same evaluation pipeline.

\section{Results}

\subsection{Main results}
\label{sec:main}

Table~\ref{tab:main-single} reports single-hop accuracy. \ours{} reaches 46.6
average EM, which is 9.6 points above the frozen model and 1.7 points above the
weight-space method LatentSkill, while shipping 9.25 times fewer parameters per
skill. Every dataset improves, with the largest absolute gains on Natural
Questions and PopQA. Among methods that compile a skill into a distributable
artifact, \ours{} is best on all four columns. Only retrieval-augmented
generation scores higher, and it does so by re-reading retrieved context at
every step, so it ships no artifact at all.

Table~\ref{tab:main-multi} reports multi-hop accuracy for the scheduled policy,
and Table~\ref{tab:ablation} reports the static and generation-only policies
beside it. Holding the shift through the whole trajectory is weak, because it
keeps the model in the short-answer mode that long chains of reasoning need to
leave. Withdrawing the shift after the first retrieval step, which we call
sched1, raises the sample-weighted multi-hop score to 30.0, above every reported
baseline row and 8.5 points above the frozen model. The improvement is broad,
with 2WikiMultihopQA, MuSiQue, and Bamboogle all moving up. The same schedule
applied to single-hop questions is neutral to slightly negative, a contrast we
return to in Section~\ref{sec:schedule}.

Table~\ref{tab:aggregate} gives the total over all seven datasets. \ours{}
reaches 37.9 EM, 9.8 points above the frozen model and 2.3 points above
LatentSkill. The best configuration drops per-skill routing and reaches 38.3
with one bundle.

Two patterns hold across method families. The first is that accuracy does not
track parameter count. Full SFT and Shared LoRA touch far more parameters than a
per-skill adapter, yet both sit below \ours{} on the single-hop average, and the
per-skill adapter sits below its own shared variant. The second is that a family
tends to have a weak side. Prompting and retrieval methods do well on multi-hop
questions, where the retrieved context supplies the intermediate facts, and
poorly on single-hop questions, where a long prompt encourages the verbose
completion that the task does not need. Fine-tuning methods show the opposite
profile, strong on single-hop questions and weak on the long chains. \ours{} is
competitive on both, which is what the behaviour account in
Section~\ref{sec:behaviour} predicts, since the gain is a change of action and
not a change of knowledge.

\begin{table*}[t]
	\centering
	\caption{Single-hop retrieval QA, grouped by method family. \textbf{Bold} marks
		the best value in a column and \underline{underline} the second best; ties share a
		mark. A shaded row marks the best method in each family. In the parameter column,
		bold marks the smallest shipped artifact, and prompt-based methods ship none.
		$\Delta$Avg is the change in Avg with respect to Vanilla, and ``Extra input''
		marks whether a method needs more than the question at inference.}
	\label{tab:main-single}
	\small
	\setlength{\tabcolsep}{11pt}
	\renewcommand{\arraystretch}{1.05}
	\begin{tabular}{@{}l ccc c c r c@{}}
		\toprule
		& \multicolumn{3}{c}{\textbf{Exact match per dataset (\%)}} & \multicolumn{3}{c}{\textbf{Summary}} & \\
		\cmidrule(lr){2-4}\cmidrule(lr){5-7}
		\textbf{Method} & \textbf{NQ}& \textbf{TriviaQA} & \textbf{PopQA} & \textbf{Avg}$\uparrow$ & \textbf{$\Delta$Avg} & \textbf{Params / skill} & \textbf{Extra input} \\
		\midrule
		\multicolumn{8}{@{}l}{\itshape Prompting, retrieval}\\
		Vanilla (base)   & 25.2 & 50.6 & 35.2 & 37.0 & \textit{ref.} & --- & \no \\
		CoT              & 19.2 & 50.8 & 20.0 & 30.0 & $-7.0$ & --- & \yes \\
		Few-Shot         & 34.6 & 57.6 & 39.4 & 43.9 & $+6.9$ & --- & \yes \\
		R1-Instruct      & 27.0 & 55.0 & 31.6 & 37.9 & $+0.9$ & --- & \yes \\
		\rowcolor{gray!25}
		RAG              & \textbf{39.0} & \textbf{64.0} & \textbf{45.0} & \textbf{49.3} & \textbf{$+12.3$} & --- & \yes \\
		In-Context Skill & 27.2 & 56.4 & 33.0 & 38.9 & $+1.9$ & 0 (in prompt) & \yes \\
		\midrule
		\multicolumn{8}{@{}l}{\itshape Fine-tuning}\\
		\rowcolor{gray!25}
		Full SFT         & 35.4 & 58.0 & 39.8 & 44.4 & $+7.4$ & all params & \no \\
		Shared LoRA      & 35.4 & 55.0 & 40.6 & 43.7 & $+6.7$ & --- & \no \\
		Per-skill LoRA   & 32.8 & 56.6 & 39.0 & 42.8 & $+5.8$ & $\sim$21.8\,M & \no \\
		\midrule
		\multicolumn{8}{@{}l}{\itshape Hypernetwork}\\
		SHINE            & 34.6 & 57.4 & 35.6 & 42.5 & $+5.5$ & --- & \no \\
		Text-to-LoRA     & 32.8 & 43.2 & 25.2 & 33.7 & $-3.3$ & --- & \no \\
		LatentSkill      & 36.2 & 57.6 & 41.0 & 44.9 & $+7.9$ & $\sim$21.8\,M & \no \\
		\rowcolor{gray!25}
		\textbf{\ours{}} & \underline{37.0} & \underline{59.4} & \underline{43.4} & \underline{46.6} & \underline{$+9.6$} & \textbf{2.36\,M} & \no \\
		\bottomrule
	\end{tabular}
\end{table*}

\begin{table*}[t]
	\centering
	\caption{Multi-hop retrieval QA, grouped by method family. \textbf{Bold} marks the
		best value in a column and \underline{underline} the second best; ties share a
		mark. A shaded row marks the best method in each family. In the parameter column,
		bold marks the smallest shipped artifact. The static and generation-only
		injection policies are reported in Table~\ref{tab:ablation}. $\Delta$Avg is
		the change in Avg with respect to Vanilla, $^{\ddagger}$ marks sample-weighted
		averages, and ``Extra input'' marks whether a method needs more than the question
		at inference.}
	\label{tab:main-multi}
	\small
	\setlength{\tabcolsep}{6.5pt}
	\renewcommand{\arraystretch}{1.05}
	\begin{tabular}{@{}l cccc c c r c@{}}
		\toprule
		& \multicolumn{4}{c}{\textbf{Exact match per dataset (\%)}} & \multicolumn{3}{c}{\textbf{Summary}} & \\
		\cmidrule(lr){2-5}\cmidrule(lr){6-8}
		\textbf{Method} & \textbf{HotpotQA}& \textbf{2Wiki} & \textbf{MuSiQue}& \textbf{Bamboogle} & \textbf{Avg}$\uparrow$ & \textbf{$\Delta$Avg} & \textbf{Params / skill} & \textbf{Extra input} \\
		\midrule
		\multicolumn{9}{@{}l}{\itshape Prompting, retrieval}\\
		Vanilla (base)   & 26.2 & 26.8 & 4.2  & 28.8 & 21.5 & \textit{ref.} & --- & \no \\
		CoT              & 22.4 & 26.2 & 5.0  & \underline{37.6} & 22.8 & $+1.3$ & --- & \yes \\
		Few-Shot         & 30.0 & 25.8 & 6.0  & 17.6 & 19.9 & $-1.6$ & --- & \yes \\
		R1-Instruct      & 26.6 & 33.6 & 5.8  & 34.4 & 25.1 & $+3.6$ & --- & \yes \\
		RAG              & 32.4 & 21.2 & 6.8  & 27.2 & 21.9 & $+0.4$ & --- & \yes \\
		\rowcolor{gray!25}
		In-Context Skill & 30.2 & \textbf{39.8} & 7.6 & \textbf{38.4} & \underline{29.0} & \underline{$+7.5$} & 0 (in prompt) & \yes \\
		\midrule
		\multicolumn{9}{@{}l}{\itshape Fine-tuning}\\
		\rowcolor{gray!25}
		Full SFT         & \underline{38.4} & 25.8 & 10.2 & 24.8 & 24.8 & $+3.3$ & all params & \no \\
		Shared LoRA      & 38.0 & 27.2 & 9.2  & 22.4 & 24.2 & $+2.7$ & --- & \no \\
		Per-skill LoRA   & 37.6 & 24.0 & 10.4 & 16.8 & 22.2 & $+0.7$ & $\sim$21.8\,M & \no \\
		\midrule
		\multicolumn{9}{@{}l}{\itshape Hypernetwork}\\
		SHINE            & 36.2 & 29.4 & \textbf{12.4} & 30.4 & 27.1 & $+5.6$ & --- & \no \\
		Text-to-LoRA     & 30.2 & 30.4 & 8.0  & 12.8 & 20.4 & $-1.1$ & --- & \no \\
		LatentSkill      & \textbf{39.6} & 32.0 & 9.8 & 25.6 & 26.8 & $+5.3$ & $\sim$21.8\,M & \no \\
		\rowcolor{gray!25}
		\textbf{\ours{}} & \underline{38.4} & \underline{38.6} & \underline{10.8} & \textbf{38.4} & \textbf{30.0}$^{\ddagger}$ & \textbf{$+8.5$} & \textbf{2.36\,M} & \no \\
		\bottomrule
	\end{tabular}
\end{table*}

\begin{table}[t]
\centering
\small
\setlength{\tabcolsep}{3pt}
\caption{All seven datasets, sample weighted. Every variant ships one small
bundle, and the best one needs no routing.}
\label{tab:aggregate}
\begin{tabular}{lcccc}
\toprule
\textbf{Method} & \textbf{1-hop} & \textbf{Multi} & \textbf{All-7} & \textbf{Params} \\
\midrule
Vanilla (base)            & 37.0 & 21.5 & 28.1 & -- \\
LatentSkill               & 44.9 & 27.0 & 35.6 & 21.8\,M \\
\midrule
Ours (static / static)    & 46.6 & 25.7 & 35.7 & \textbf{2.36\,M} \\
Ours (static / sched1)    & 46.6 & 30.0 & 37.9 & \textbf{2.36\,M} \\
Ours (6000 steps)         & 43.7 & 30.0 & 36.6 & \textbf{2.36\,M} \\
Ours (universal)          & 47.1 & 30.2 & \textbf{38.3} & \textbf{2.36\,M} \\
\bottomrule
\end{tabular}
\end{table}

\subsection{The shift shapes behaviour}
\label{sec:behaviour}

\begin{figure*}[t]
\centering
\includegraphics[width=0.245\textwidth]{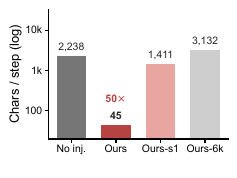}\hfill
\includegraphics[width=0.245\textwidth]{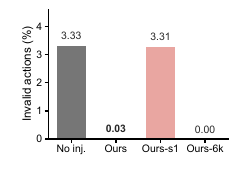}\hfill
\includegraphics[width=0.245\textwidth]{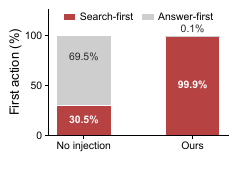}\hfill
\includegraphics[width=0.245\textwidth]{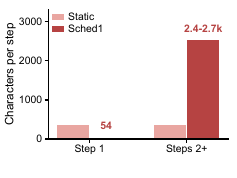}
\caption{What the compiled shift does. It shortens each step, removes invalid
actions, flips the first action to search, and, under the schedule, keeps long
reasoning after the first step.}
\label{fig:behaviour}
\end{figure*}

The gain is not extra knowledge. It is a change of action. Table~\ref{tab:behavior}
and Figure~\ref{fig:behaviour} show what the compiled shift does to a single-hop
trajectory. Without injection the model writes an average of 2238 characters per
step and produces an invalid action in 3.33\% of steps. With the shift the
average falls to 45 characters and invalid actions to 0.03\%. The first action
flips from answer-first in 30.5\% of questions to search-first in 99.9\%. The
model stops deliberating in prose and starts retrieving, then answering in a few
words. That switch is the direct source of the single-hop gain.

The same view explains why the multi-hop schedule works. Under a static shift
the model stays short at every step, which suppresses the reasoning that a
multi-hop question needs. Under sched1 the first step still triggers retrieval
in about 54 characters, and later steps return to long reasoning of 2400 to 2700
characters. The schedule separates the retrieval trigger from the reasoning
that follows.

\begin{table}[t]
\centering
\small
\setlength{\tabcolsep}{3pt}
\caption{Behaviour of the compiled shift. The no-injection rows come from the
same pipeline as the reported runs.}
\label{tab:behavior}
\begin{tabular}{lcccc}
\toprule
\textbf{Setting} & \textbf{EM} & \textbf{Steps} & \textbf{Inv.\%} & \textbf{Chars} \\
\midrule
\multicolumn{5}{l}{\emph{Single-hop (1500)}} \\
\quad No injection & 38.5 & 1.38 & 3.33 & 2238 \\
\quad \textbf{\ours{}} & \textbf{46.6} & 2.15 & \textbf{0.03} & \textbf{45} \\
\quad \ours{} (sched1) & 45.8 & 2.16 & 3.31 & 1411 \\
\midrule
\multicolumn{5}{l}{\emph{Multi-hop (1625)}} \\
\quad No injection & 25.2 & 1.65 & 5.77 & 3287 \\
\quad \textbf{\ours{} (sched1)} & \textbf{30.0} & 2.59 & 6.00 & 2515 \\
\quad \ours{} (static) & 25.7 & 3.02 & -- & 373 \\
\bottomrule
\end{tabular}
\end{table}

\subsection{What the shift injects}
\label{sec:inject}

\begin{figure}[t]
\centering
\includegraphics[width=0.49\linewidth]{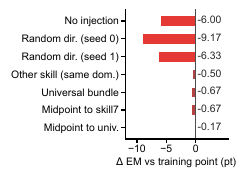}\hfill
\includegraphics[width=0.49\linewidth]{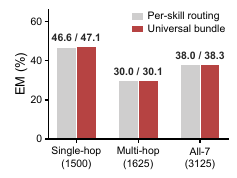}
\caption{Direction controls and the universal bundle. Replacing the trained
component with a same-norm random offset falls back to the no-injection level,
while one averaged bundle matches per-skill routing.}
\label{fig:prior}
\end{figure}

We now test what the compiled shift carries. Table~\ref{tab:routing} gathers
three controls, and Figure~\ref{fig:prior} shows the direction control.

\paragraph{A wrong skill is as good as the right one.}
Replacing the skill document with one from another task domain leaves single-hop
accuracy unchanged, at 47.9 against 46.6 for the correct skill. A wrong skill
from the same domain gives 46.3. If the artifact carried per-skill content, a
wrong document should hurt. It does not.

\paragraph{One bundle replaces all skills.}
Averaging the per-layer factors of several skills into a single bundle, and
deploying that bundle with no routing, reaches 47.1 on single-hop and 30.2 on
multi-hop, at or above per-skill routing. A two-sided equivalence test on paired
question ids cannot reject equivalence within two points for either pool. The
practical reading is that one 2.36M bundle covers the whole benchmark.

\paragraph{A random direction does not.}
The two results above could suggest that any shift works. It does not. When we
remove the trained component and add a random offset of the same norm, accuracy
falls to the no-injection level, from 46.5 to 37.3 and 40.2 across two seeds,
while the no-injection control sits at 40.5. Adding random noise on top of the
trained direction, by contrast, costs almost nothing. The direction is
load-bearing, and it has to come from training.

\paragraph{Weight-space methods behave the same way.}
A universal LoRA adapter, obtained by averaging per-skill adapters, matches
per-skill LoRA on single-hop questions, and a LoRA built from a wrong skill is
again unchanged. The equivalence between a universal artifact and a per-skill
artifact therefore belongs to the task and the behavioural prior, not to the
activation-space parameterisation.

Taken together, the four observations describe the carrier as a domain-general
behavioural prior. Any in-distribution skill document reaches it, which is why
routing can be dropped. The document still has a role. It selects which mode is
triggered, and that role becomes visible when a family needs more than one mode,
as the next subsection shows.

\begin{table*}[t]
\centering
\small
\setlength{\tabcolsep}{3pt}
\caption{What the shift carries. A wrong skill and a universal bundle match
per-skill routing, a random direction does not, and weight-space adapters
follow the same pattern.}
\label{tab:routing}
\begin{minipage}[t]{0.42\textwidth}
\centering
\textbf{(a) Skill routing (single-hop)}\\[2pt]
\begin{tabular}{lcccc}
\toprule
\textbf{injected content} & \textbf{NQ} & \textbf{Triv.} & \textbf{PopQA} & \textbf{Avg} \\
\midrule
Correct skill        & 37.0 & 59.4 & 43.4 & 46.6 \\
Wrong (cross-dom.)   & 39.4 & 61.0 & 43.2 & \textbf{47.9} \\
Wrong (same-dom.)    & 37.0 & 58.4 & 43.4 & 46.3 \\
\bottomrule
\end{tabular}
\end{minipage}\hfill
\begin{minipage}[t]{0.52\textwidth}
\centering
\textbf{(b) Universal bundle (no routing)}\\[2pt]
\begin{tabular}{lccc}
\toprule
\textbf{configuration} & \textbf{Single} & \textbf{Multi} & \textbf{All-7} \\
\midrule
Per-skill routing            & 46.60 & 29.97 & 37.95 \\
\textbf{Universal (9 skills)} & \textbf{47.13} & \textbf{30.15} & \textbf{38.30} \\
Universal (3 skills)         & 46.33 & 30.58 & 38.14 \\
\bottomrule
\end{tabular}
\end{minipage}

\vspace{6pt}

\textbf{(c) Paired equivalence tests (two-sided exact McNemar; TOST $\pm$2\,pt)}\\[2pt]
\begin{tabular}{llcccc}
\toprule
\textbf{contrast} & \textbf{pool} & \textbf{$n$} & \textbf{$\Delta$ EM (pt)} & \textbf{$p$} & \textbf{equivalent} \\
\midrule
Universal (9) vs per-skill routing      & single-hop & 1500 & $+0.53$ & 0.4752 & \textbf{yes} \\
Universal (9) vs universal (3)          & single-hop & 1500 & $+0.80$ & 0.2513 & \textbf{yes} \\
Universal (9)+sched1 vs routing+sched1  & multi-hop  & 1625 & $+0.18$ & 0.8732 & \textbf{yes} \\
Universal (9)+sched1 vs universal (9)+static & multi-hop & 1625 & $+6.58$ & $<10^{-4}$ & no \\
\bottomrule
\end{tabular}
\end{table*}

\subsection{Schedulability and routing-free deployment}
\label{sec:schedule}

The behaviour view gives a criterion for when the shift should be withdrawn. If
a task's gain can be split into a short trigger and a longer process, then
removing the shift after the trigger frees the process. Multi-hop questions have
this shape, and sched1 adds 4.2 points over the static shift, significant under
a paired test. Single-hop questions do not, because most of them close within
two steps, and removing the shift there costs 0.8 points. The criterion predicts
both signs.

Deployment follows from the same finding. Since one bundle is equivalent to
per-skill routing, a server can ship a single 2.36M artifact and skip the router
entirely. This cuts the stored parameters from nine bundles to one and removes a
per-query routing decision, with no loss in accuracy and a small gain on the
aggregate.

The recipe has three consequences for a serving stack. Storage scales with the
number of task families and not the number of skills, so adding a skill to a
family costs nothing at deployment. Routing disappears, which removes both the
label lookup and its failure mode, since a mislabelled query can no longer reach
the wrong artifact. The bundle also loads in one step, so a warm server holds a
single low-rank adapter per family in place of a directory of per-skill files.
The cost is a small loss of control, because one bundle cannot be tuned to a
single task family. For the families we test, that control is not worth keeping,
and the aggregate result in Table~\ref{tab:aggregate} shows the trade favours
the single bundle.

\subsection{Ablations and robustness}

\paragraph{Strength is a threshold.}
Scaling the shift below its trained value does not weaken the behaviour
gradually. At 0.25 and 0.5 the model still writes long completions and accuracy
stays near the no-injection level. Only the trained strength of 1.0 triggers the
short search-then-answer mode, and 1.5 breaks the model entirely. The shift
encodes a discrete mode that is switched on, not a dial.

\paragraph{One ninth of the compute is enough.}
Stage two is best at 2400 micro-steps. Training longer raises per-step length
and gives the single-hop gain back, which matches the behaviour account, since
more training drifts the model toward its original verbose form. The reported
numbers therefore come from about one ninth of the reference compute.

\paragraph{Injection phase is tied to the protocol.}
Because SearchQA drives actions from the prompt context with reasoning disabled,
the shift has to cover the prompt forward pass as well as generation. Applying
it only during generation leaves the prompt unshaped and collapses accuracy. The
phase is a property of the evaluation protocol, not a free hyperparameter.

\paragraph{The artifact has a wide usable neighbourhood.}
Perturbing the injected content along random directions by up to 300\% leaves
accuracy within about one point of the trained value, whereas a weight-space
adapter collapses once the perturbation passes about a third of its norm. The
usable radius is roughly an order of magnitude wider. The plateau is wide in
direction and narrow in magnitude, which is the signature of a mode, not a
strength.

\begin{figure}[t]
\centering
\includegraphics[width=0.49\linewidth]{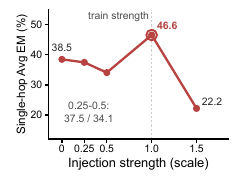}\hfill
\includegraphics[width=0.49\linewidth]{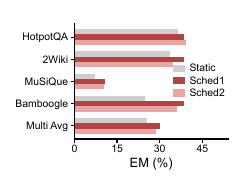}
\caption{Strength acts as a threshold, not a dial, and the per-step
schedule improves multi-hop accuracy while the single-hop control stays flat.}
\label{fig:ablation}
\end{figure}

\begin{figure}[t]
\centering
\includegraphics[width=0.49\linewidth]{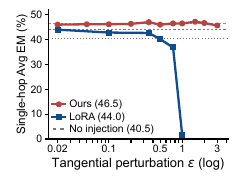}\hfill
\includegraphics[width=0.49\linewidth]{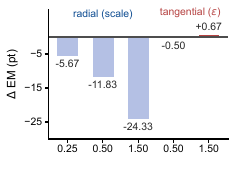}
\caption{The compiled shift keeps its accuracy under large random perturbations,
while a weight-space adapter collapses. The plateau is wide in direction and
narrow in magnitude.}
\label{fig:robust}
\end{figure}

\section{Discussion}

The controls point to a single account. A hypernetwork trained across many agent
skills learns a domain-general behavioural prior, and a skill document is a key
that unlocks it. The teeth of the key matter less than the distribution it comes
from, since any in-distribution document reaches the same prior while a random
direction does not. What the model gains is a way of acting, such as retrieving
before answering, and not a store of facts about the task.

The domain-generality of the prior has a simple source. The generator is trained
across many skill documents, and the signal that shapes it is an action trace
and not a document. A document that never appears in the trace is still a valid
conditioning input, so the generator learns to map a broad range of
in-distribution documents to the same action mode. The document acts as a
switch, and the switch is coarse by construction. This reading fits the two
controls that pull in opposite directions. The trained direction is
load-bearing, while the specific document is not, so a random direction fails
even when a wrong document succeeds.

This account reframes several choices that look like design details. Routing
becomes optional, because one artifact reaches the prior. Per-skill storage
becomes unnecessary, because averaging preserves the prior. The schedule becomes
a tool for releasing the prior once it has done its work. It also explains why
the shift is a threshold, not a dial. A mode is on or off, so an
intermediate strength either fails to switch or overdrives the model.

The same account applies to weight-space compilation. A universal LoRA adapter
matches per-skill adapters, and a wrong-skill adapter is unchanged, so the
equivalence is a property of the task and the prior, and not of the
activation space. The advantage of the activation space is then one of cost. A
low-rank post-block offset is 9.25 times smaller per skill, and its usable
neighbourhood is about an order of magnitude wider, which makes it a lighter and
more robust carrier for the same prior.

\section{Conclusion}

We asked what a compiled skill injection carries. On a retrieval agent, three
controls show that the carrier is a domain-general behavioural prior, not the
content of the skill. We built \ours{} on this finding, a one-pass
hypernetwork that maps a skill document to low-rank activation shifts at every
block output. The artifact is 9.25 times smaller than a weight-space adapter,
beats it on a seven-dataset retrieval benchmark, can be withdrawn mid-trajectory
to release long reasoning, and can be shipped as a single routing-free bundle.

\section*{Limitations}

Our evidence comes from a single task family, retrieval question answering, and
a single base model. The behavioural prior we identify is the retrieve-then-answer
policy that this family rewards. Tasks whose skills differ in factual content
and not in policy may lean more on the document, and we do not claim that
the prior dominates there. We also evaluate one base model, so the prior is
measured relative to that model and its training distribution. Finally, our
training corpus of agent skills is fixed, and we do not study how the prior
changes when the corpus shifts.

\bibliography{skill2steer}

@misc{li2026latentharnesscompilingagent,
	title={Latent Harness: Compiling Agent Behaviors into Post-Block Activation Shifts}, 
	author={Ran Li and Shimin Di and Zeyu Ren and Ling Yue and Jiachuan Wang and Lei Chen},
	year={2026},
	eprint={2611.01415},
	archivePrefix={arXiv},
	primaryClass={cs.CL},
	url={https://arxiv.org/abs/2611.01415}, 
}

@article{kwiatkowski2019natural,
	title={Natural questions: a benchmark for question answering research},
	author={Kwiatkowski, Tom and Palomaki, Jennimaria and Redfield, Olivia and Collins, Michael and Parikh, Ankur and Alberti, Chris and Epstein, Danielle and Polosukhin, Illia and Devlin, Jacob and Lee, Kenton and others},
	journal={Transactions of the Association for Computational Linguistics},
	volume={7},
	pages={453--466},
	year={2019},
	publisher={MIT Press One Rogers Street, Cambridge, MA 02142-1209, USA journals-info~…}
}

@inproceedings{yang2018hotpotqa,
	title={HotpotQA: A dataset for diverse, explainable multi-hop question answering},
	author={Yang, Zhilin and Qi, Peng and Zhang, Saizheng and Bengio, Yoshua and Cohen, William and Salakhutdinov, Ruslan and Manning, Christopher D},
	booktitle={Proceedings of the 2018 conference on empirical methods in natural language processing},
	pages={2369--2380},
	year={2018}
}

@inproceedings{ho2020constructing,
	title={Constructing a multi-hop qa dataset for comprehensive evaluation of reasoning steps},
	author={Ho, Xanh and Nguyen, Anh-Khoa Duong and Sugawara, Saku and Aizawa, Akiko},
	booktitle={Proceedings of the 28th International Conference on Computational Linguistics},
	pages={6609--6625},
	year={2020}
}

@inproceedings{joshi2017triviaqa,
	title={Triviaqa: A large scale distantly supervised challenge dataset for reading comprehension},
	author={Joshi, Mandar and Choi, Eunsol and Weld, Daniel S and Zettlemoyer, Luke},
	booktitle={Proceedings of the 55th Annual Meeting of the Association for Computational Linguistics (Volume 1: Long Papers)},
	pages={1601--1611},
	year={2017}
}

@inproceedings{mallen2023not,
	title={When not to trust language models: Investigating effectiveness of parametric and non-parametric memories},
	author={Mallen, Alex and Asai, Akari and Zhong, Victor and Das, Rajarshi and Khashabi, Daniel and Hajishirzi, Hannaneh},
	booktitle={Proceedings of the 61st annual meeting of the association for computational linguistics (volume 1: Long papers)},
	pages={9802--9822},
	year={2023}
}

@article{trivedi2022musique,
	title={♫ musique: Multihop questions via single-hop question composition},
	author={Trivedi, Harsh and Balasubramanian, Niranjan and Khot, Tushar and Sabharwal, Ashish},
	journal={Transactions of the Association for Computational Linguistics},
	volume={10},
	pages={539--554},
	year={2022},
	publisher={MIT Press One Broadway, 12th Floor, Cambridge, Massachusetts 02142, USA~…}
}

@inproceedings{press2023measuring,
	title={Measuring and narrowing the compositionality gap in language models},
	author={Press, Ofir and Zhang, Muru and Min, Sewon and Schmidt, Ludwig and Smith, Noah A and Lewis, Mike},
	booktitle={Findings of the Association for Computational Linguistics: EMNLP 2023},
	pages={5687--5711},
	year={2023}
}

@article{yang2025qwen3,
	title={Qwen3 technical report},
	author={Yang, An and Li, Anfeng and Yang, Baosong and Zhang, Beichen and Hui, Binyuan and Zheng, Bo and Yu, Bowen and Gao, Chang and Huang, Chengen and Lv, Chenxu and others},
	journal={arXiv preprint arXiv:2505.09388},
	year={2025}
}

@article{wang2024multilingual,
	title={Multilingual e5 text embeddings: A technical report},
	author={Wang, Liang and Yang, Nan and Huang, Xiaolong and Yang, Linjun and Majumder, Rangan and Wei, Furu},
	journal={arXiv preprint arXiv:2402.05672},
	year={2024}
}

@article{lu2026skill0,
	title={Skill0: In-context agentic reinforcement learning for skill internalization},
	author={Lu, Zhengxi and Yao, Zhiyuan and Wu, Jinyang and Han, Chengcheng and Gu, Qi and Cai, Xunliang and Lu, Weiming and Xiao, Jun and Zhuang, Yueting and Shen, Yongliang},
	journal={arXiv preprint arXiv:2604.02268},
	year={2026}
}

@article{li2026skillsinjector,
	title={SkillsInjector: Dynamic Skill Context Construction for LLM Agents},
	author={Li, Yanchao and Liu, Wanhao and Gao, Ben and Xie, Jiaqing and Ai, Zhehong and Zou, Na and Li, Yuqiang and Fu, Tianfan},
	journal={arXiv preprint arXiv:2605.29794},
	year={2026}
}

@article{yu2026latentskill,
	title={Latentskill: From in-context textual skills to in-weight latent skills for llm agents},
	author={Yu, Aofan and Zhou, Chenyu and Xu, Tianyi and Guo, Zihan and Shan, Rong and Fu, Zhihui and Wang, Jun and Liu, Weiwen and Yu, Yong and Zhang, Weinan and others},
	journal={arXiv preprint arXiv:2606.06087},
	year={2026}
}

@article{zhao2026parametric,
	title={Parametric Skills},
	author={Zhao, Xuan and He, Haonan and Yang, Qingyu and Li, Minglei and Ye, Jingqi and Tan, Zelin and Wan, Bo and Ye, Peng},
	journal={arXiv preprint arXiv:2606.30015},
	year={2026}
}

@article{liu2026shine,
	title={Shine: A scalable in-context hypernetwork for mapping context to lora in a single pass},
	author={Liu, Yewei and Wang, Xiyuan and Mao, Yansheng and Gelbery, Yoav and Maron, Haggai and Zhang, Muhan},
	journal={arXiv preprint arXiv:2602.06358},
	year={2026}
}

@article{charakorn2025text,
	title={Text-to-lora: Instant transformer adaption},
	author={Charakorn, Rujikorn and Cetin, Edoardo and Tang, Yujin and Lange, Robert Tjarko},
	journal={arXiv preprint arXiv:2506.06105},
	year={2025}
}

@article{charakorn2026doc,
	title={Doc-to-lora: Learning to instantly internalize contexts},
	author={Charakorn, Rujikorn and Cetin, Edoardo and Uesaka, Shinnosuke and Lange, Robert Tjarko},
	journal={arXiv preprint arXiv:2602.15902},
	year={2026}
}

@article{sun2025hypersteer,
	title={Hypersteer: Activation steering at scale with hypernetworks},
	author={Sun, Jiuding and Baskaran, Sidharth and Wu, Zhengxuan and Sklar, Michael and Potts, Christopher and Geiger, Atticus},
	journal={arXiv preprint arXiv:2506.03292},
	year={2025}
}

@inproceedings{sharma2026cold,
	title={COLD-steer: Steering large language models via in-context one-step learning dynamics},
	author={Sharma, Kartik and Trivedi, Rakshit},
	booktitle={International Conference on Learning Representations},
	volume={2026},
	pages={27736--27756},
	year={2026}
}

@article{zhang2026skill,
	title={Skill-to-LoRA: From Using Skills to Learning Behaviors for Token-Efficient LLM Agents},
	author={Zhang, Tianyi and Qi, Zhonghao},
	journal={arXiv preprint arXiv:2606.16769},
	year={2026}
}

@inproceedings{shenaj2025lora,
	title={Lora. rar: Learning to merge loras via hypernetworks for subject-style conditioned image generation},
	author={Shenaj, Donald and Bohdal, Ondrej and Ozay, Mete and Zanuttigh, Pietro and Michieli, Umberto},
	booktitle={2025 IEEE/CVF International Conference on Computer Vision (ICCV)},
	pages={16132--16142},
	year={2025},
	organization={IEEE}
}

@inproceedings{ye2021learning,
	title={Learning to generate task-specific adapters from task description},
	author={Ye, Qinyuan and Ren, Xiang},
	booktitle={Proceedings of the 59th Annual Meeting of the Association for Computational Linguistics and the 11th International Joint Conference on Natural Language Processing (Volume 2: Short Papers)},
	pages={646--653},
	year={2021}
}

@inproceedings{tan2026instant,
	title={Instant personalized large language model adaptation via hypernetwork},
	author={Tan, Zhaoxuan and Zhang, Zixuan and Wen, Haoyang and Li, Zheng and Zhang, Rongzhi and Chen, Pei and Mo, Fengran and Liu, Zheyuan and Zeng, Qingkai and Yin, Qingyu and others},
	booktitle={Proceedings of the 64th Annual Meeting of the Association for Computational Linguistics (Volume 1: Long Papers)},
	pages={23557--23580},
	year={2026}
}

@article{adila2026weight,
	title={Weight Updates as Activation Shifts: A Principled Framework for Steering},
	author={Adila, Dyah and Cooper, John and Yun, Alexander and Trost, Avi and Sala, Frederic},
	journal={arXiv preprint arXiv:2603.00425},
	year={2026}
}

@inproceedings{rimsky2024steering,
	title={Steering llama 2 via contrastive activation addition},
	author={Rimsky, Nina and Gabrieli, Nick and Schulz, Julian and Tong, Meg and Hubinger, Evan and Turner, Alexander},
	booktitle={Proceedings of the 62nd Annual Meeting of the Association for Computational Linguistics (Volume 1: Long Papers)},
	pages={15504--15522},
	year={2024}
}

@article{zou2023representation,
	title={Representation engineering: A top-down approach to ai transparency},
	author={Zou, Andy and Phan, Long and Chen, Sarah and Campbell, James and Guo, Phillip and Ren, Richard and Pan, Alexander and Yin, Xuwang and Mazeika, Mantas and Dombrowski, Ann-Kathrin and others},
	journal={arXiv preprint arXiv:2310.01405},
	year={2023}
}

@article{wu2024reft,
	title={Reft: Representation finetuning for language models},
	author={Wu, Zhengxuan and Arora, Aryaman and Wang, Zheng and Geiger, Atticus and Jurafsky, Dan and Manning, Christopher D and Potts, Christopher},
	journal={Advances in Neural Information Processing Systems},
	volume={37},
	pages={63908--63962},
	year={2024}
}

@article{turner2023steering,
	title={Steering language models with activation engineering},
	author={Turner, Alexander Matt and Thiergart, Lisa and Leech, Gavin and Udell, David and Vazquez, Juan J and Mini, Ulisse and MacDiarmid, Monte},
	journal={arXiv preprint arXiv:2308.10248},
	year={2023}
}

@article{li2023inference,
	title={Inference-time intervention: Eliciting truthful answers from a language model},
	author={Li, Kenneth and Patel, Oam and Vi{\'e}gas, Fernanda and Pfister, Hanspeter and Wattenberg, Martin},
	journal={Advances in neural information processing systems},
	volume={36},
	pages={41451--41530},
	year={2023}
}

@inproceedings{todd2024function,
	title={Function vectors in large language models},
	author={Todd, Eric and Li, Millicent and Sen Sharma, Arnab and Mueller, Aaron and Wallace, Byron and Bau, David},
	booktitle={International conference on learning representations},
	volume={2024},
	pages={17282--17333},
	year={2024}
}

@misc{li2026learningdecidereasonparameterefficient,
	title={Learning to Decide, Not to Reason: Parameter-Efficient Decision Operators via Low-Rank Activation Steering}, 
	author={Ran Li and Lei Chen},
	year={2026},
	eprint={2610.06950},
	archivePrefix={arXiv},
	primaryClass={cs.LG},
	url={https://arxiv.org/abs/2610.06950}, 
}

\appendix

\section{Appendix}

\subsection{Ablation details}
\label{app:ablation}

Table~\ref{tab:ablation} collects the four sweeps behind
Section~\ref{sec:behaviour}.

\paragraph{Injection strength.}
The scale sweep has one valid operating point. At scale 0.00 the no-injection
control reaches 38.5 average EM and writes 2238 characters per step. Scaling to
0.25 and 0.50 leaves the model in its verbose mode, at 2247 and 2315 characters
per step, and accuracy stays at 37.5 and 34.1. A partial strength does not
weaken the behaviour, it fails to trigger it. At the trained strength of 1.00
accuracy rises to 46.6 and per-step length collapses to 45 characters. At 1.50
the model overdrives, writing 5790 characters per step and falling to 22.2. The
sweep is a threshold, and the trained value is the only point that both switches
the mode and keeps it stable.

\paragraph{Stage-two training budget.}
The best single-hop checkpoint is 2400 micro-steps at 46.60, and a rerun of the
same checkpoint gives 46.33, which places the run-to-run spread near 0.3 points.
Training further lowers single-hop accuracy to 42.00 at 3200 steps and 40.00 at
4000 steps, while per-step length climbs back to 1309 and 1630 characters. At
6000 steps the bundle recovers to 43.67 and holds multi-hop at 30.03. The
pattern matches the behaviour account, since more training drifts the model
toward its original verbose form and undoes the short action mode. Every
reported result therefore comes from about one ninth of the reference compute.

\paragraph{Injection phase.}
Covering only the generation phase drops single-hop accuracy from 46.6 to 23.9
and multi-hop accuracy from 25.7 to 14.2, while per-step length rises to 3697 and
5108 characters. SearchQA drives actions from the prompt context with reasoning
disabled, so the prompt forward pass has to be shaped as well. The phase is a
property of the protocol and not a free hyperparameter.

\paragraph{Per-step schedule.}
Static injection reaches 25.7 on multi-hop questions. Withdrawing the shift
after step 1, which we call sched1, raises the score to 30.0, and holding it for
two steps, sched2, gives 28.8. The gain of sched1 is largest on Bamboogle, from
24.8 to 38.4, and on 2WikiMultihopQA, from 33.6 to 38.6. The single-hop control
moves the other way, from 46.6 under static injection to 45.8 under sched1,
because most single-hop questions close within two steps and there is no long
reasoning left to release.

\begin{table*}[t]
\centering
\small
\setlength{\tabcolsep}{3pt}
\caption{Ablations over injection strength, training budget, injection phase,
and per-step schedule.}
\label{tab:ablation}
\begin{minipage}[t]{0.49\textwidth}
\centering
\textbf{(a) Injection strength (scale)}\\[2pt]
\begin{tabular}{lccccc}
\toprule
\textbf{scale} & \textbf{NQ} & \textbf{Triv.} & \textbf{PopQA} & \textbf{Avg} & \textbf{Chars} \\
\midrule
0.00 & 28.2 & 56.4 & 31.0 & 38.5 & 2238 \\
0.25 & 26.4 & 57.2 & 29.0 & 37.5 & 2247 \\
0.50 & 22.4 & 52.0 & 27.8 & 34.1 & 2315 \\
\textbf{1.00} & \textbf{37.0} & \textbf{59.4} & \textbf{43.4} & \textbf{46.6} & \textbf{45} \\
1.50 & 16.2 & 35.0 & 15.4 & 22.2 & 5790 \\
\bottomrule
\end{tabular}
\end{minipage}\hfill
\begin{minipage}[t]{0.49\textwidth}
\centering
\textbf{(b) Stage-2 training budget}\\[2pt]
\begin{tabular}{lccc}
\toprule
\textbf{micro-steps} & \textbf{Single-hop} & \textbf{Multi-hop} & \textbf{Chars} \\
\midrule
\textbf{2400} & \textbf{46.60} & \textbf{29.97} & \textbf{45} \\
2400 (rerun)  & 46.33 & -- & 47 \\
3200          & 42.00 & -- & 1309 \\
4000          & 40.00 & -- & 1630 \\
6000          & 43.67 & 30.03 & 3132 \\
\bottomrule
\end{tabular}
\end{minipage}

\vspace{6pt}

\begin{minipage}[t]{0.49\textwidth}
\centering
\textbf{(c) Injection phase}\\[2pt]
\begin{tabular}{lccc}
\toprule
\textbf{phase} & \textbf{Single} & \textbf{Multi} & \textbf{Chars} \\
\midrule
\textbf{both (full)} & \textbf{46.6} & 25.7 & 45 / 373 \\
gen-only & 23.9 & 14.2 & 3697 / 5108 \\
\bottomrule
\end{tabular}
\end{minipage}\hfill
\begin{minipage}[t]{0.49\textwidth}
\centering
\textbf{(d) Per-step injection schedule}\\[2pt]
\begin{tabular}{lccccc}
\toprule
\textbf{schedule} & \textbf{Hotpot} & \textbf{2Wiki} & \textbf{MuSiQ.} & \textbf{Bamboo} & \textbf{Multi} \\
\midrule
static   & 36.6 & 33.6 & 7.2  & 24.8 & 25.7 \\
gen-only & 17.2 & 21.2 & 2.0  & 23.2 & 14.2 \\
\textbf{sched1} & \textbf{38.4} & \textbf{38.6} & \textbf{10.8} & \textbf{38.4} & \textbf{30.0} \\
sched2   & 39.4 & 34.8 & 10.4 & 36.0 & 28.8 \\
\bottomrule
\end{tabular}
\end{minipage}
\end{table*}

\subsection{Robustness of the compiled artifact}
\label{app:robust}

Table~\ref{tab:robust} reports the perturbation study on a shared set of 600
single-hop questions. Every configuration runs on the same question ids and
through the same evaluation script as the main results.

\paragraph{Tangential scan.}
Adding noise along random directions keeps \ours{} within 0.83 points of the
training point at every $\varepsilon$ up to 3.0, which changes the injected
content by up to 207\%. Per-step length stays in the short mode across the whole
range, from 44 to 409 characters. A weight-space adapter behaves differently. It
loses 1.33 points at $\varepsilon = 0.35$, 7.17 points at 0.75, and collapses to
1.50 EM at 1.0. The usable neighbourhood of the activation shift is about an
order of magnitude wider than that of the weight-space adapter. This is a
property of the artifact and not of the evaluation, since both are measured on
the same ids with the same script.

\paragraph{Direction controls.}
Replacing the trained component with a same-norm random offset drops accuracy to
37.33 and 40.17 across two seeds, at or below the no-injection control at 40.50.
Swapping in another skill from the same domain costs 0.50 points, a universal
bundle costs 0.67, and two interpolations toward other training points cost 0.67
and 0.17. The trained direction is necessary, while the specific training point
is not. The contrast between these two rows is the central evidence for the
behaviour-prior account.

\paragraph{Radial against tangential.}
The plateau around the trained point is wide in direction and narrow in
magnitude. Scaling the shift radially to 0.25 and 0.50 costs 5.67 and 11.83
points, and 1.50 costs 24.33 points. Tangential perturbations of comparable size
cost under one point. A mode that is switched on has this shape, because the
direction decides which mode is entered while the magnitude decides whether the
mode is entered at all.

\begin{table*}[t]
\centering
\small
\setlength{\tabcolsep}{4pt}
\caption{Robustness of the compiled artifact on a shared question set.}
\label{tab:robust}
\begin{minipage}[t]{0.52\textwidth}
\centering
\textbf{(a) Tangential scan $\varepsilon$}\\[2pt]
\begin{tabular}{lcccccc}
\toprule
\textbf{$\varepsilon$} & 0 & 0.02 & 0.10 & 0.35 & 0.50 & 0.75 \\
\midrule
Ours        & 46.50 & 46.00 & 46.17 & 47.00 & 46.00 & 46.50 \\
Weight LoRA & 44.00 & 44.00 & 42.83 & 42.67 & 40.17 & 36.83 \\
\midrule
\textbf{$\varepsilon$} & 1.00 & 1.50 & 2.00 & 3.00 & & \\
\midrule
Ours        & 46.50 & 47.17 & 46.67 & 45.67 & & \\
Weight LoRA & {1.50} & -- & -- & -- & & \\
\bottomrule
\end{tabular}
\end{minipage}\hfill
\begin{minipage}[t]{0.45\textwidth}
\centering
\textbf{(b) Direction controls (single-hop)}\\[2pt]
\begin{tabular}{lcc}
\toprule
\textbf{perturbation} & \textbf{EM} & \textbf{$\Delta$ EM (pt)} \\
\midrule
No injection                   & 40.50 & $-6.00$ \\
Random dir. (seed 0)           & 37.33 & $-9.17$ \\
Random dir. (seed 1)           & 40.17 & $-6.33$ \\
Other skill (same dom.)        & 46.00 & $-0.50$ \\
Universal bundle               & 45.83 & $-0.67$ \\
Midpoint to skill7             & 45.83 & $-0.67$ \\
Midpoint to universal          & 46.33 & $-0.17$ \\
\bottomrule
\end{tabular}
\end{minipage}
\end{table*}

\subsection{Routing, universal bundles, and equivalence}
\label{app:routing}

Table~\ref{tab:routing} reports the routing study behind
Section~\ref{sec:inject}.

\paragraph{Skill routing.}
On single-hop questions the correct skill gives 46.6, a cross-domain wrong skill
gives 47.9, and a same-domain wrong skill gives 46.3. The spread across the
three is smaller than the run-to-run variation reported above, so the document
does not carry per-skill content that the model can use.

\paragraph{Universal bundle.}
Averaging the per-layer factors of nine skills into one bundle reaches 47.13 on
single-hop questions and 30.15 on multi-hop questions, at or above per-skill
routing at 46.60 and 29.97. A bundle averaged over the three SearchQA skills
reaches 46.33 and 30.58. Raising the retrieval budget from top-3 to top-5 leaves
the single-hop result at 47.53, while lowering it to top-1 falls to 43.27, so
top-3 is close to saturated and the default budget is a reasonable choice.

\paragraph{Equivalence.}
A two-sided exact McNemar test with a margin of two points finds the universal
bundle equivalent to per-skill routing on single-hop questions at a difference
of 0.53 points, and on multi-hop questions at a difference of 0.18 points. The
same test separates the schedule from the static policy at 6.58 points and from
the two-step policy at 4.68 points, and it separates the top-1 retrieval budget
from the default at 3.87 points. Equivalence therefore holds where the
differences are small, and significance holds where the schedule or the
retrieval budget actually changes behaviour.

\subsection{Behaviour statistics for longer budgets}
\label{app:behaviour}

Table~\ref{tab:behavior-detail} restates the behaviour statistics for the two
settings that Section~\ref{sec:behaviour} does not cover. At 6000 training steps
the bundle reaches 43.7 single-hop EM at 1.93 steps per question and 3132
characters per step, close to the no-injection length of 2238, which shows the
return to the verbose mode. The generation-only policy on multi-hop questions
gives 14.2 EM at 5108 characters per step.

\begin{table}[t]
\centering
\small
\setlength{\tabcolsep}{3pt}
\caption{Behaviour statistics for the longer-budget variants.}
\label{tab:behavior-detail}
\begin{tabular}{lcccc}
\toprule
\textbf{Setting} & \textbf{EM} & \textbf{Steps} & \textbf{Inv.\%} & \textbf{Chars} \\
\midrule
Ours (6000 steps) & 43.7 & 1.93 & 0.00 & 3132 \\
Ours (gen-only, multi) & 14.2 & 1.01 & -- & 5108 \\
\bottomrule
\end{tabular}
\end{table}

\subsection{Training and export details}
\label{app:training}

For completeness we list the settings that produce the reported bundles. The
base model is Qwen3-8B with 36 decoder blocks and a hidden width of 4096. The
skill encoder appends 16 memory tokens and runs a single frozen forward pass.
The generator is a four-layer transformer encoder with width 4096, 32 heads, and
an 8192-wide feed-forward network, and it holds about 537.3M parameters that are
used only at compile time. Stage one trains on about 170K skill documents under
the reconstruction and completion objectives, and the skill text enters only the
encoder. Stage two trains on about 8500 skill and trace pairs for 2400
micro-steps on a single A800, with the injection strength fixed at $10^{-2}$ and
the rank fixed at 8. The loss is the cross-entropy on expert action tokens, and
the base model stays frozen throughout. At export the strength is folded into
the up factor, so a deployed bundle is a set of 36 low-rank factor pairs, or
2.36M scalars. The bundle mounts through the native steering interface of the
inference engine with an explicit scale, and the engine and the base model are
never modified.

\section{LLM usage}
\label{app:llm}
Large-language-model tools were used for editing and formatting support during manuscript preparation.

\end{document}